%% file: neurips_2026.tex
\documentclass{article}

\usepackage[dblblindworkshop,final]{neurips_2026}
\workshoptitle{2nd Workshop on Agentic AI Benchmarks and Applications for Enterprise Tasks}

\usepackage[utf8]{inputenc}
\usepackage[T1]{fontenc}
\usepackage{graphicx}
\usepackage{booktabs}
\usepackage{microtype}
\usepackage{xcolor}
\usepackage{hyperref}
\usepackage{url}
\usepackage{enumitem}
\usepackage{float}
\hypersetup{hidelinks}
\usepackage{booktabs}
\usepackage{tabularx}
\usepackage{longtable}

\newcommand{\pb}{\textsc{PointerBench}}

\title{PointerBench: A Diagnostic Benchmark for Fine-Grained GUI Grounding}

\author{%
  \textbf{Valentino Sacco}$^{1,2}$\thanks{Work conducted during an internship at Warmwind AG.}, \ 
  \textbf{Saurabh Pandey}$^{2}$, \ 
  \textbf{Alexander Florian Jercher}$^{2}$, \ 
  \textbf{Maximilian Schilling}$^{2}$ \\[3pt]
  $^{1}$Sapienza University of Rome \qquad $^{2}$Warmwind AG
}

\begin{document}
\maketitle

\input{sec/0_abstract}
\input{sec/1_introduction}

\begin{figure}[t]
  \centering
  \includegraphics[width=\linewidth]{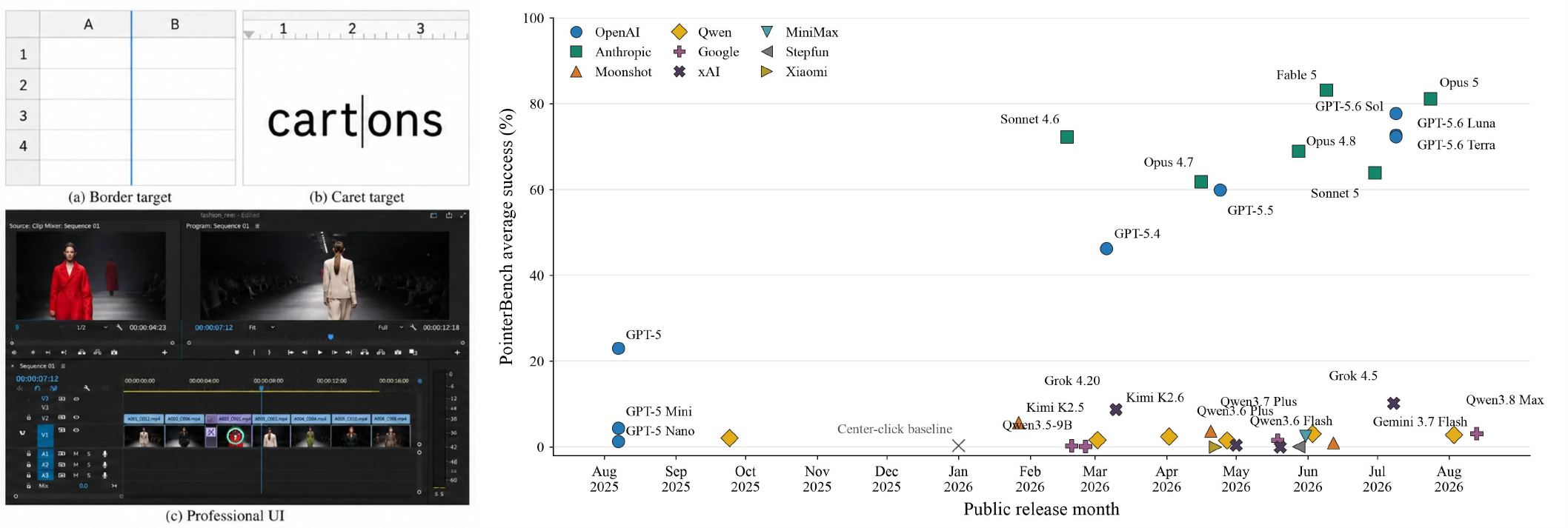}
  \vspace{-4mm}
  \caption{PointerBench evaluates fine-grained grounding across spreadsheet geometry, character-level text targets, and professional interfaces. Results reveal substantial headroom and pronounced differences across model families.
}
  \label{fig:teaser}
\end{figure}

\input{sec/2_related_work}
\input{sec/3_analysis}
\input{sec/5_conclusion}

\clearpage
\bibliographystyle{plainnat}
\bibliography{references}

\appendix
\input{sec/a_supplementary}

\end{document}

%% file: sec/0_abstract.tex
\begin{abstract}
Multimodal Large Language Models have shown promising performance in computer-use tasks, from planning user workflows to grounding natural-language actions to precise UI targets.
Existing end-to-end benchmarks primarily
emphasize long-horizon task completion, while GUI grounding benchmarks remain
focused on conventional controls and a limited set of professional interfaces. We
introduce \pb{}, a diagnostic suite of 1,500 screenshot--instruction
pairs probing three complementary capabilities: thin sub-element identification
and geometric grounding in spreadsheets (\pb{}-Sheets), character-level
grounding in long-text documents (\pb{}-Text), and mixed GUI
grounding across 100 professional applications (\pb{}-Pro). We carefully
curate each task with pixel-exact target geometry and evaluate both general-purpose
multimodal models and 39 specialist GUI grounding configurations. We find text grounding to be
the most challenging setting, with Claude Fable 5, GPT-5.6 Sol, and the leading
native specialist UI-Venus-2-9B achieving $69.6\%$, $53.4\%$, and $25.8\%$
success on \pb{}-Text despite respective $83.2\%$, $77.7\%$, and $54.5\%$
average success across \pb{}. These results reveal substantial
headroom in fine-grained GUI grounding across frontier and specialist models. We publicly release the benchmark and evaluation code at \url{https://github.com/warmwindOS/pointerbench}.
\end{abstract}

%% file: sec/1_introduction.tex
\section{Introduction}


Computer-use agents have emerged as promising day-to-day assistants, automating workflows across browsers, desktop applications, mobile devices, and specialized professional software~\citep{deng2023mind2web,zhou2024webarena,xie2024osworld,rawles2025androidworld,li2025screenspotpro}. Multimodal agents interleave reasoning and action~\citep{yao2023react}, combining long-horizon planning~\citep{erdogan2025planandact}, external tool use~\citep{schick2023toolformer,qin2024toolllm}, and visual grounding~\citep{cheng2024seeclick,gou2025uground}. For tasks such as ``open the meeting chat'' or ``paste cell B14 to cell E15'', successful execution requires both selecting an appropriate action sequence and grounding each step to precise screen coordinates or regions~\citep{zheng2024seeact,gou2025uground}. This remains challenging in modern graphical user interfaces (GUIs), which often contain dense and visually ambiguous interaction targets, as well as in high-resolution professional applications with small target regions~\citep{li2025screenspotpro,wu2025guiactor}.

Existing evaluations measure complementary but incomplete aspects of this problem. Interactive benchmarks such as Online-Mind2Web, WebArena, and OSWorld evaluate long-horizon task completion in websites or executable desktop environments~\citep{xue2025an,zhou2024webarena,xie2024osworld}. They are essential for studying planning and recovery, but end-to-end success conflates failures in reasoning, grounding, execution, and environment control. Static benchmarks such as ScreenSpot and ScreenSpot-Pro isolate localization and have accelerated progress on visual GUI agents~\citep{cheng2024seeclick,li2025screenspotpro}. Yet they lack fine-grained actions such as placing a cursor on a spreadsheet boundary, between two characters, or around a narrowly specified text span. Rapid progress on these benchmarks motivates complementary tests that vary the \emph{geometry} of the requested action, diagnosing an agent's ability to perform precise control.

We introduce \pb{}, a fine-grained diagnostic benchmark comprising three 500-example tasks designed to diagnose these missing capabilities (Figure~\ref{fig:teaser}). \textbf{\pb{}-Sheets} probes cells, headers, colors, relative references, edges, resize handles, and corners under varied sheet layouts. \textbf{\pb{}-Text} evaluates points and bounding boxes over words, characters, punctuation, carets, and invoice fields in six languages. \textbf{\pb{}-Pro} covers icon, text, and mixed targets across 100 professional applications. All inputs use a fixed $1024\!\times\!768$ coordinate frame and ship with executable scorers.

Our contributions are twofold: (i) a diagnostic benchmark suite that isolates structural, textual, and professional GUI grounding through three complementary tasks with precisely defined target geometry; and (ii) a comprehensive evaluation spanning both general-purpose multimodal models and specialist GUI grounding configurations, revealing non-trivial rank reversals across subsets and a persistent bottleneck in character-level and region-level textual grounding.

%% file: sec/2_related_work.tex
\section{Related Work}

\noindent\textbf{Interactive computer-use benchmarks.}
Mind2Web contains over 2,000 open-ended web tasks across 137 websites and records the corresponding action sequences~\citep{deng2023mind2web}. WebArena provides reproducible, functional websites for evaluating long-horizon task execution~\citep{zhou2024webarena}, while OSWorld extends this paradigm to real web and desktop applications with execution-based evaluators~\citep{xie2024osworld}. AndroidWorld similarly evaluates autonomous agents over real mobile applications~\citep{rawles2025androidworld}. Recent work further studies the separation between high-level planning and low-level execution in long-horizon agents~\citep{erdogan2025planandact}. These benchmarks test complete interactive agents, whereas \pb{} deliberately removes temporal planning to isolate whether an already-specified action can be grounded precisely.

\paragraph{GUI grounding.}
ScreenSpot introduced screenshot-only grounding across mobile, desktop, and
web interfaces~\citep{cheng2024seeclick}, while ScreenSpot-Pro extends this
setting to dense, high-resolution professional applications~\citep{li2025screenspotpro}.
OSWorld-G further decomposes computer-use grounding into recognition, text
matching, layout reasoning, and precise manipulation~\citep{xie2025osworldg}.
UI-Vision provides fine-grained offline evaluation across 83 desktop
applications through element grounding, layout grounding, and action
prediction~\citep{nayak2025uivision}, while FineState-Bench evaluates
state-conditioned grounding and exact GUI state setting across desktop, web,
and mobile interfaces~\citep{ji2026finestate}. 
Alongside these benchmarks, specialized GUI agents~\citep{qin2025uitars,wang2025opencua,liu2026scalecua,zhou2025mai} and grounding models~\citep{gou2025uground,gu2025uivenus,venusteam2026uivenus15,uivenus2026,chen2026uiins} have substantially improved localization on conventional controls. Complementary advances include grounding architectures~\citep{wu2025guiactor,ouyang2026focusui,jiang2026rexomni}, reinforcement learning methods~\citep{yang2025gta1,tang2026guig2,liu2026infiguig1,yuan2025segui}, and inference-time prediction aggregation~\citep{zhang2026mvp,du2026guirc}.
\pb{} complements
these efforts by systematically isolating specific sub-element geometries
under controlled, pixel-exact generation: thin spreadsheet boundaries,
character- and caret-level text locations, and tightly bounded regions, while
\pb{}-Pro broadens coverage to 100 professional interfaces. This provides a
reproducible diagnostic of grounding errors independently of temporal planning
and environment execution.

%% file: sec/3_analysis.tex
\section{PointerBench}
\label{sec:benchmark}

\noindent\textbf{Task design.}
In interactive environments, a grounding error may move an agent into an
unintended state and propagate through subsequent actions
~\citep{xie2024osworld,xie2025osworldg}.
Consider, for example, a spreadsheet task requiring an agent to resize a column
or select a cell of a particular color. Although simple to specify in language,
these actions require different forms of fine-grained grounding: the former
requires localizing a thin boundary between adjacent columns, whereas the latter
requires identifying a visually specified region within a repetitive grid.

\pb{} isolates these capabilities from long-horizon planning through three
complementary 500-example suites: \pb{}-Sheets targets structural and geometric
grounding, \pb{}-Text targets fine-grained textual grounding, and \pb{}-Pro
evaluates grounding across heterogeneous professional interfaces.
Each benchmark instance consists of a $1024\!\times\!768$ screenshot, a
natural-language instruction, and a target expressed in absolute pixel
coordinates. Depending on the task, the required output is either a point or a
bounding box. Point predictions are successful when they fall inside the
annotated target region. Bounding-box predictions require at least $0.90$
target coverage and $0.70$ prediction precision; malformed outputs are counted
as failures.
We adopt a standardized $1024\!\times\!768$ resolution for two reasons: first,
fine-grained spatial capabilities, such as localizing 1--3\,px spreadsheet
dividers, inter-character carets, and individual glyphs, become substantially
harder at higher resolutions as relative target area shrinks; second, higher
resolutions markedly increase 
the time taken by a model to reason. Downscaling to $1024\!\times\!768$ makes
evaluation computationally efficient while isolating sub-element geometric
precision.
Dataset construction and reproducibility details are provided in the
Appendix.

\vspace{-4mm}
\input{tables/main_table}

\vspace{-4mm}
\subsection{Frontier Model Performance}
\label{sec:results}

Table~\ref{tab:main} reports a representative subset of our evaluation, selected
to span six model families, multiple generations, and a broad range of
grounding capabilities. All models are evaluated with reasoning. 
An analysis of sensitivity to the reasoning budget and the complete evaluation spanning more than 30
multimodal models (see Fig.~\ref{fig:teaser}) are reported in the Appendix.

Claude Fable 5 leads \pb{} with an average success rate of $83.2\%$, followed
by Claude Opus 5 at $81.1\%$ and GPT-5.6 Sol at $77.7\%$. GPT-5.6 Luna,
GPT-5.6 Terra, and Claude Sonnet 4.6 follow at $72.7\%$, $72.3\%$, and
$72.2\%$, respectively. Performance,
however, is strongly task dependent. Opus leads \pb{}-Sheets at $98.8\%$,
narrowly ahead of Sol and Fable at $97.8\%$ and $97.6\%$. The three models
are close on \pb{}-Pro, reaching $81.2\%$, $82.0\%$, and $82.4\%$,
respectively, while \pb{}-Text separates them more clearly: Fable reaches
$69.6\%$, compared with $63.2\%$ for Opus and $53.4\%$ for Sol. At the lower end, 
Kimi K2.5 reaches only $5.7\%$. Gemini 3.7 Flash and
Qwen3.8 Max obtain official scores of $3.1\%$ and $2.8\%$, respectively,
but these results are dominated by systematic coordinate-frame
non-compliance and are analyzed separately in Sec.~\ref{sec:per-suite}.

\input{tables/specialist_main_table}

\vspace{-3mm}
\subsection{Specialist GUI Grounding Models}
\label{sec:specialists}
\vspace{-1mm}

Because dedicated GUI agents are trained under model-specific coordinate conventions,
we evaluate 39 open-weight specialist GUI grounding configurations (2B--32B parameters)
using each model's native grounding prompt and parser (Table~\ref{tab:specialist-main};
full 39-run leaderboard in Appendix~\ref{app:specialists}). Across the method families in
Table~\ref{tab:specialist-main}, Multi-View Prediction (MVP), which aggregates predictions
across multiple augmented views at inference time, is the leading UI specialist,
scoring best on every track with its 32B backbone
($68.4\%$ Sheets, $34.0\%$ Text, $75.4\%$ Pro, $59.3\%$ Avg.).
Behind MVP-32B, a rank reversal occurs across suites among trailing models:
UI-Venus-2-9B (pairing domain-specific grounding supervision with reinforcement learning)
performs second best on Pro ($73.6\%$, surpassing all single-view 32B models),
while GTA1-32B (an RL-trained grounding model that directly predicts click coordinates)
does so on Text ($29.6\%$) and overall ($54.9\%$), and OpenCUA-32B
(trained on large-scale multi-OS trajectories) on Sheets ($66.0\%$).
Considering smaller models ($<9$B), MAI-UI-8B (an instruction-tuned desktop/mobile grounding agent)
performs best at $51.3\%$ average, followed closely by MVP-8B ($51.2\%$),
UI-Venus-1.5-8B ($51.0\%$), and GTA1-7B ($50.1\%$), ahead of OpenCUA-7B ($47.3\%$),
UI-Venus-Ground-7B ($46.2\%$), and UI-TARS-1.5-7B ($40.6\%$).

\subsection{Per-Suite Analysis}
\label{sec:per-suite}

\noindent\textbf{\pb{}-Sheets.}
\pb{}-Sheets isolates the spatial structure of spreadsheet interfaces, where
successful interaction may depend on reasoning over a grid rather than simply
recognizing a conventional GUI element. The suite includes cells, headers,
color-based targets, relative references, thin boundaries and resize handles,
and cell corners, separating conventional element localization from structural
and geometric grounding. Sheets is the strongest suite for the leading models,
with Opus, Sol, and Fable reaching $98.8\%$, $97.8\%$, and $97.6\%$,
respectively. Conventional cell and header targets are localized almost
perfectly, while performance drops on finer geometric interactions: Fable
reaches $78.1\%$ on row-resize handles, showing that near-perfect aggregate
performance can still conceal errors on thin structural elements.

\noindent\textbf{\pb{}-Text.}
\pb{}-Text probes fine-grained textual grounding through words, characters,
punctuation, caret positions, and tightly specified regions. Targets may occupy
only a narrow portion of a line or lie between adjacent glyphs, with point and
bounding-box tasks spanning six languages and multiple document surfaces.

\input{tables/text_table}

\begin{figure}[t]
  \centering
  \includegraphics[width=0.90\linewidth]{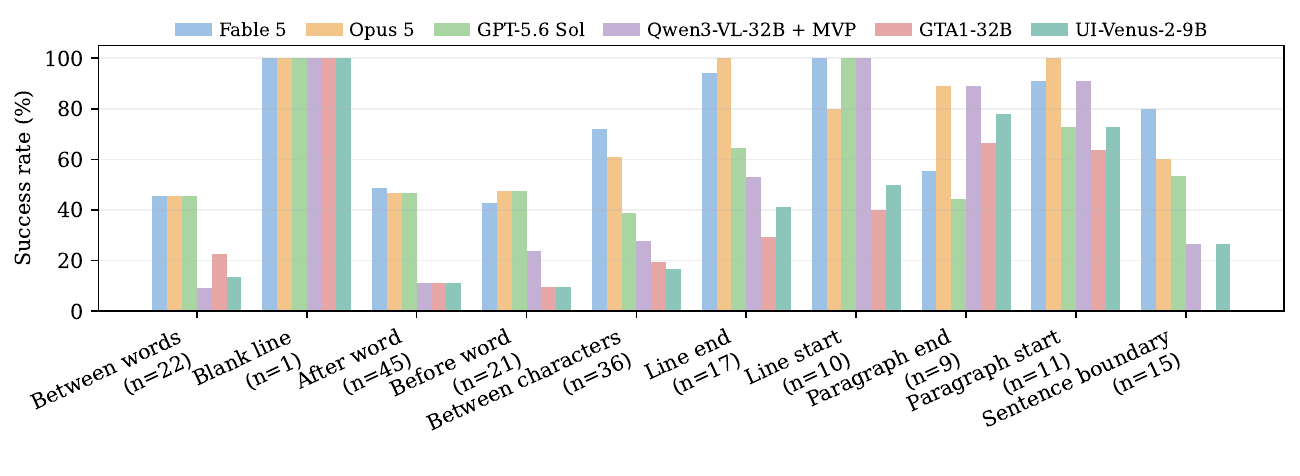}
  \vspace{-3mm}
  \caption{\pb{}-Text caret accuracy by placement objective.}
  \vspace{-4mm}
  \label{fig:caret-study}
\end{figure}

Table~\ref{tab:text-medium-target-groups} shows a strong dependence on target
granularity across both frontier and specialist systems.
Despite strong conventional localization (e.g., UI-Venus-2-9B reaches
$89.0\%$--$100.0\%$ on spreadsheet cells/headers, vs.\ $31.6\%$--$55.8\%$
on borders/resize handles and $3.0\%$ on relative offsets), \pb{}-Text
is the primary bottleneck across all specialist configurations, with no specialist
exceeding $34.0\%$ (MVP-32B $34.0\%$, GTA1-32B $29.6\%$, UI-Venus-2-9B $25.8\%$,
MAI-UI-8B $23.2\%$). Word localization is easiest: Fable reaches
$90.2\%$, followed by Opus ($83.9\%$), Sol ($76.8\%$), and top specialists
GTA1-32B ($70.5\%$) and MVP-32B ($69.6\%$). Accuracy drops
sharply at finer granularity: Opus and Sol reach $51.9\%$/$48.1\%$ on
characters and $39.1\%$/$56.5\%$ on punctuation, while top UI specialists fall
to $11.1$--$18.5\%$ on characters and $22.5$--$34.2\%$ on carets ($19.3\%$ for MAI-UI-8B).
Region grounding, requiring
tight bounds over text spans or fields, remains challenging for frontier models
(Fable $56.0\%$, Opus $50.7\%$). Specialists whose native output is a point are scored
on that point and therefore receive $0.0\%$ on all 134 region tasks.
When a model returns a box, that box is scored: GUI-G2-7B~\citep{tang2026guig2} reaches $12.7\%$ on Region,
and UI-Venus-Ground-7B reaches $11.2\%$ ($24.6\%$ overall on \pb{}-Text).
Figure~\ref{fig:caret-study} further decomposes caret grounding: placements
immediately before, after, or between words are non-trivial, with
no frontier model exceeding $50\%$ accuracy and UI specialists remaining below $24\%$,
whereas coarse structural targets (line start, paragraph start/end) are localized more reliably.

\noindent\textbf{\pb{}-Pro.}
\pb{}-Pro evaluates grounding across visually heterogeneous professional
interfaces. It spans 100 applications targeting icons, text elements, and
interface controls such as dropdowns, list items, sliders, tabs, and panels.
Performance is consistent among the leading models, with Fable,
Opus, Sol, and Luna reaching $82.4\%$, $81.2\%$, $82.0\%$, and $79.4\%$,
respectively. Text controls are easiest, followed by other interface controls,
while icons are hardest. Icons account for $57.3\%$ of pooled failures despite
representing only $38.0\%$ of examples, indicating that a major remaining
challenge lies in mapping an intended action to the correct visual control and
highlighting icon understanding and UI semantics as important targets for
improving computer-use agents.

\noindent\textbf{Coordinate-frame compliance.}
Gemini 3.7 Flash and Qwen3.8 Max predominantly return coordinates in a
$0$--$1000$ frame despite being explicitly instructed to use absolute
$1024\times768$ coordinates; the exact prompt is reported in the Appendix.
Their official scores are therefore $3.1\%$ and $2.8\%$, respectively.
Uniformly converting parseable predictions to the requested frame raises these
to $78.1\%$ and $55.4\%$, placing Gemini within the performance range of the
benchmark's leading models. This suggests that grounding performance can depend
strongly on model-native coordinate conventions and motivates evaluation under
normalized output representations.

%% file: tables/main_table.tex
\begin{table*}[h]
\centering
\caption{Success rate (\%) on the three \textsc{PointerBench} subsets. Best and second-best results are \textbf{bold} and \underline{underlined}, respectively.}
\label{tab:main}
\resizebox{\textwidth}{!}{
\begin{tabular}{lcccccccccc}
\toprule
& \multicolumn{3}{c}{Claude}
& \multicolumn{3}{c}{GPT}
& Grok & Kimi & Gemini & Qwen \\
\cmidrule(lr){2-4}
\cmidrule(lr){5-7}
\cmidrule(lr){8-8}
\cmidrule(lr){9-9}
\cmidrule(lr){10-10}
\cmidrule(lr){11-11}
&
Sonnet 4.6 &
Fable 5 &
Opus 5 &
5.6 Luna &
5.6 Terra &
5.6 Sol &
4.5 &
K2.5 &
3.7 Flash &
3.8 Max \\
\midrule
Sheets
& 88.8 & 97.6 & \textbf{98.8}
& 92.0 & 91.6 & \underline{97.8}
& 15.2 & 10.8 & 4.0 & 4.8 \\
Text
& 49.0 & \textbf{69.6} & \underline{63.2}
& 46.6 & 47.6 & 53.4
& 1.4 & 3.4 & 1.2 & 0.2 \\
Pro
& 78.8 & \textbf{82.4} & 81.2
& 79.4 & 77.6 & \underline{82.0}
& 14.0 & 3.0 & 4.2 & 3.4 \\
\midrule
Avg.
& 72.2 & \textbf{83.2} & \underline{81.1}
& 72.7 & 72.3 & 77.7
& 10.2 & 5.7 & 3.1 & 2.8 \\
\bottomrule
\end{tabular}
\vspace{-3mm}
}
\end{table*}

%% file: tables/specialist_main_table.tex
\begin{table*}[t]
\centering
\caption{Success rate (\%) of specialist GUI grounding configurations on the three \textsc{PointerBench} subsets (full 39-run evaluation in Appendix~\ref{app:specialists}). Best and second-best results are \textbf{bold} and \underline{underlined}, respectively.}
\label{tab:specialist-main}
\resizebox{\textwidth}{!}{
\begin{tabular}{lcccccccccccc}
\toprule
& UI-TARS
& \multicolumn{2}{c}{GTA1}
& \multicolumn{2}{c}{OpenCUA}
& \multicolumn{3}{c}{UI-Venus}
& \multicolumn{2}{c}{MAI-UI}
& \multicolumn{2}{c}{MVP} \\
\cmidrule(lr){2-2}
\cmidrule(lr){3-4}
\cmidrule(lr){5-6}
\cmidrule(lr){7-9}
\cmidrule(lr){10-11}
\cmidrule(lr){12-13}
&
1.5-7B &
7B &
32B &
7B &
32B &
Ground-7B &
1.5-8B &
2-9B &
2B &
8B &
8B &
32B \\
\midrule
Sheets
& 46.2
& 60.2 & 65.0
& 59.8 & \underline{66.0}
& 47.8 & 56.4 & 64.2
& 36.6 & 59.2
& 60.4 & \textbf{68.4} \\
Text
& 18.2
& 22.8 & \underline{29.6}
& 18.4 & 19.4
& 24.6 & 23.8 & 25.8
& 9.8 & 23.2
& 22.4 & \textbf{34.0} \\
Pro
& 57.4
& 67.4 & 70.2
& 63.8 & 67.4
& 66.2 & 72.8 & \underline{73.6}
& 61.0 & 71.6
& 70.8 & \textbf{75.4} \\
\midrule
Avg.
& 40.6
& 50.1 & \underline{54.9}
& 47.3 & 50.9
& 46.2 & 51.0 & 54.5
& 35.8 & 51.3
& 51.2 & \textbf{59.3} \\
\bottomrule
\end{tabular}
\vspace{-4mm}
}
\end{table*}

%% file: tables/text_table.tex
\begin{table*}[h]
\centering
\small
\caption{\pb{}-Text accuracy (\%) by target objective for the top frontier models (top) and representative UI specialist configurations (bottom). Region groups
bounding-box targets; other columns score point-coordinate
predictions. Best and second-best results are bold and underlined.}
\label{tab:text-medium-target-groups}
\begin{tabular}{lrrrrrr}
\toprule
Model & Word & Character & Punctuation & Caret & Interface text & Region \\
\midrule
Claude Fable 5
& \textbf{90.2}
& \textbf{70.4}
& \textbf{78.3}
& \textbf{64.7}
& 82.4
& \textbf{56.0} \\

Claude Opus 5
& \underline{83.9}
& \underline{51.9}
& 39.1
& \underline{62.6}
& 82.4
& \underline{50.7} \\

GPT-5.6 Sol
& 76.8
& 48.1
& \underline{56.5}
& 51.9
& \textbf{100.0}
& 30.6 \\
\midrule
Qwen3-VL-32B + MVP
& 69.6
& 18.5
& 43.5
& 34.2
& 76.5
& 0.0 \\

GTA1-32B
& 70.5
& 11.1
& 39.1
& 22.5
& \underline{88.2}
& 0.0 \\

UI-Venus-2-9B
& 52.7
& 11.1
& 26.1
& 25.7
& 76.5
& 0.0 \\

UI-Venus-Ground-7B
& 58.9
& 11.1
& 17.4
& 10.7
& \underline{88.2}
& 11.2 \\
\bottomrule
\end{tabular}
\end{table*}

%% file: sec/5_conclusion.tex
\vspace{-3mm}
\section{Conclusion}
\vspace{-2mm}
We introduced \pb{}, a diagnostic benchmark for fine-grained GUI grounding
across spreadsheet structure, text, and professional interfaces. Evaluation
across both frontier multimodal systems and specialist GUI grounding
configurations reveals that aggregate performance can hide markedly different
grounding capabilities. Leading models reliably localize conventional
spreadsheet and application controls, yet remain challenged by thin structural
elements; textual grounding degrades as interaction moves from words
to characters, carets, and precisely bounded regions; and professional-interface
failures concentrate disproportionately on icons. 
By exposing these failures
at the level of individual grounding primitives, \pb{} provides a targeted
test bed for measuring and improving precise visual control in computer-use agents.

%% file: sec/a_supplementary.tex
\section{Benchmark Construction and Reproducibility}
\label{app:data-generation}

\pb{} contains three 500-example suites with a shared interface: a
$1024\!\times\!768$ screenshot, a natural-language instruction, and pixel-exact
target geometry. Sheets and Text are generated programmatically from controlled
interface descriptions, while Pro is assembled from a larger collection of
generated professional interfaces. The released metadata stores the instruction,
target geometry, answer type, and diagnostic tags used by the scorers. We open source the benchmark metadata and evaluation script required to score it.

\paragraph{\pb{}-Sheets.}
Each example is produced by sampling a visible spreadsheet grid, layout, visual
style, and target type, and then rendering the resulting interface together with
its instruction and target coordinates. The visible sheet is varied through row
and column offsets, style changes, and optional application chrome, preventing a
fixed correspondence between semantic locations and screen coordinates. Targets
include cells and headers as well as colors, relative references, resize handles,
edges, and corners. Generation is deterministic, yielding one distinct
screenshot and metadata record per example.

\paragraph{\pb{}-Text.}
Text examples are rendered from controlled document surfaces with exact glyph
geometry. Tasks cover words, characters, punctuation, caret positions, text
spans, and structured document fields, with 366 point targets and 134
bounding-box targets. The latter include word-, line-, character-, paragraph-,
and invoice-field regions. The language mix is fixed to $50\%$ English and
$10\%$ each German, French, Spanish, Italian, and Dutch. Generation is
deterministic, and examples are retained only when they identify a unique
target.

\paragraph{\pb{}-Pro.}
Pro is assembled from generated professional-application interfaces. Candidate
targets are sampled across applications and balanced so that each of the 100
applications contributes five examples. The resulting suite contains 190 icon
targets, 154 text targets, and 156 other interface controls. For icon examples,
instructions explicitly request the icon itself to avoid ambiguity with nearby
text labels. Application name, category, platform, and source identifier are
retained in the released metadata, allowing each benchmark example to be traced
to its originating interface. The final selection is deterministic and
reproducible.

\section{Benchmark Composition and Geometry}
\label{app:composition}

Table~\ref{tab:sheets-taxonomy} summarizes the target
composition of the Sheets suite. The complete list of the 100 Pro applications
and the source identifiers of all examples are distributed with the benchmark.

\begin{table}[ht]
    \centering
    \small
    \setlength{\tabcolsep}{5pt}
    \caption{Task composition of \pb{}-Sheets. Counts sum to 500 examples.}
    \label{tab:sheets-taxonomy}
    \begin{tabularx}{\linewidth}{lXr}
        \toprule
        Family & Fine-grained targets & \# \\
        \midrule
        Cell
        & reference (82), reference with content (45), content (43)
        & 170 \\
        Edge
        & column resize (43), row resize (32), right cell edge (22),
          bottom cell edge (19)
        & 116 \\
        Relative
        & relative offset (33), relative row (27)
        & 60 \\
        Header
        & column header (35), row header (24)
        & 59 \\
        Color
        & color-specified cell or region (56)
        & 56 \\
        Corner
        & bottom-left (13), bottom-right (12), top-left (9), top-right (5)
        & 39 \\
        \midrule
        \textbf{Total} & & \textbf{500} \\
        \bottomrule
    \end{tabularx}
\end{table}

\begin{table}[t]
    \centering
    \small
    \caption{Task composition of \pb{}-Text. Region groups bounding-box
    targets; the remaining groups require point predictions.}
    \label{tab:text-taxonomy}
    \begin{tabularx}{\linewidth}{lXr}
        \toprule
        Objective & Target type & \# \\
        \midrule
        Word & word-level locations & 112 \\
        Character & individual characters & 27 \\
        Punctuation & punctuation marks & 23 \\
        Caret & insertion positions within and around text & 187 \\
        Region & text spans and structured document fields & 134 \\
        Interface text & text in document-interface controls & 17 \\
        \midrule
        \textbf{Total} & & \textbf{500} \\
        \bottomrule
    \end{tabularx}
\end{table}

\begin{figure}[t]
    \centering
    \includegraphics[width=0.86\linewidth]
        {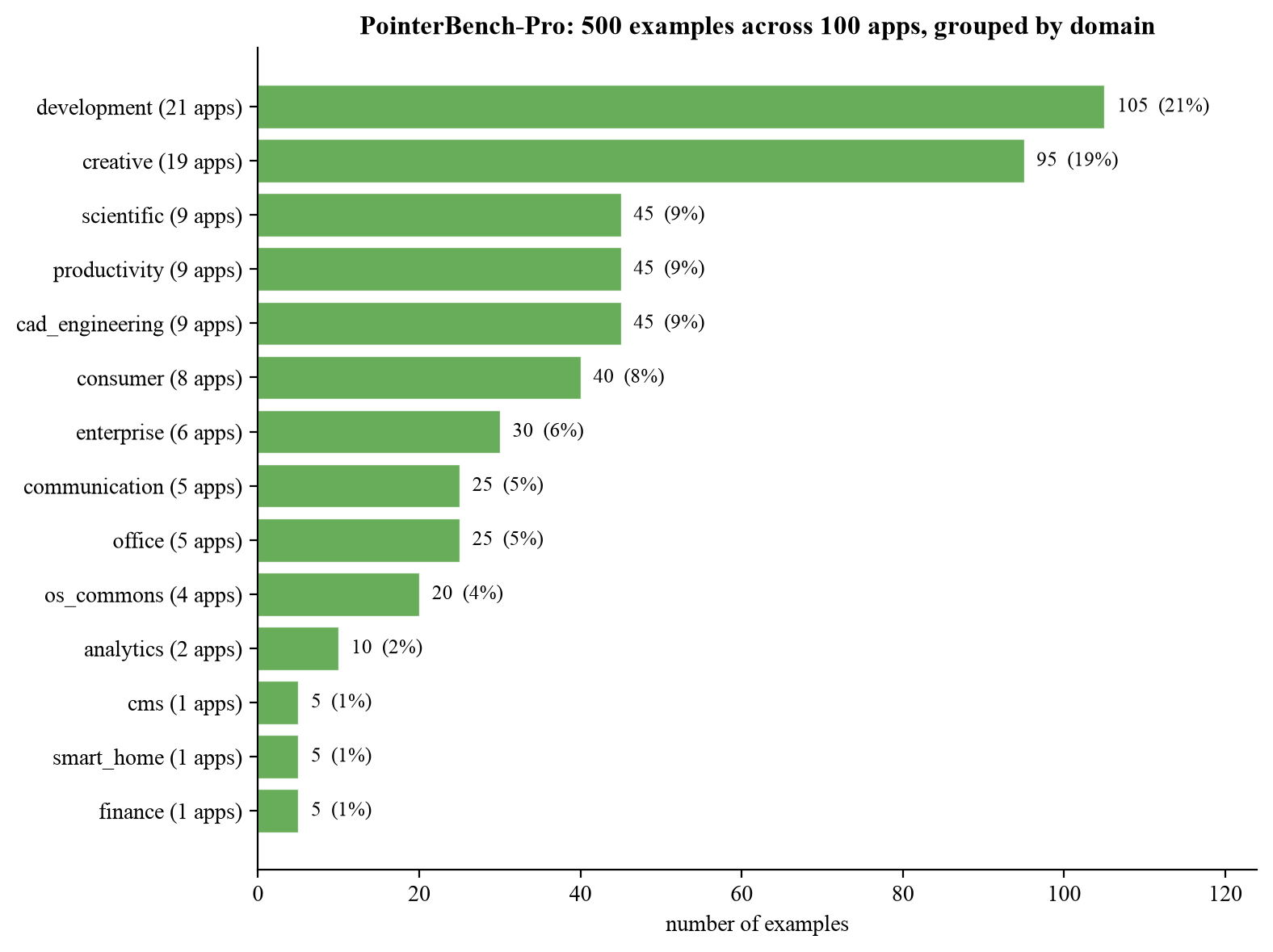}
    \caption{Application-domain coverage of \pb{}-Pro. Each of the 100
    applications contributes five examples; numbers in parentheses denote the
    number of applications in each domain.}
    \label{fig:pro-domains}
\end{figure}

\paragraph{Professional application coverage.}
Figure~\ref{fig:pro-domains} summarizes the 100 represented applications by
domain. Development and creative software provide the largest groups, while
the remaining examples span scientific, engineering, productivity,
communication, enterprise, office, operating-system, and consumer interfaces.

\begin{figure}[t]
    \centering
    \includegraphics[width=\linewidth]{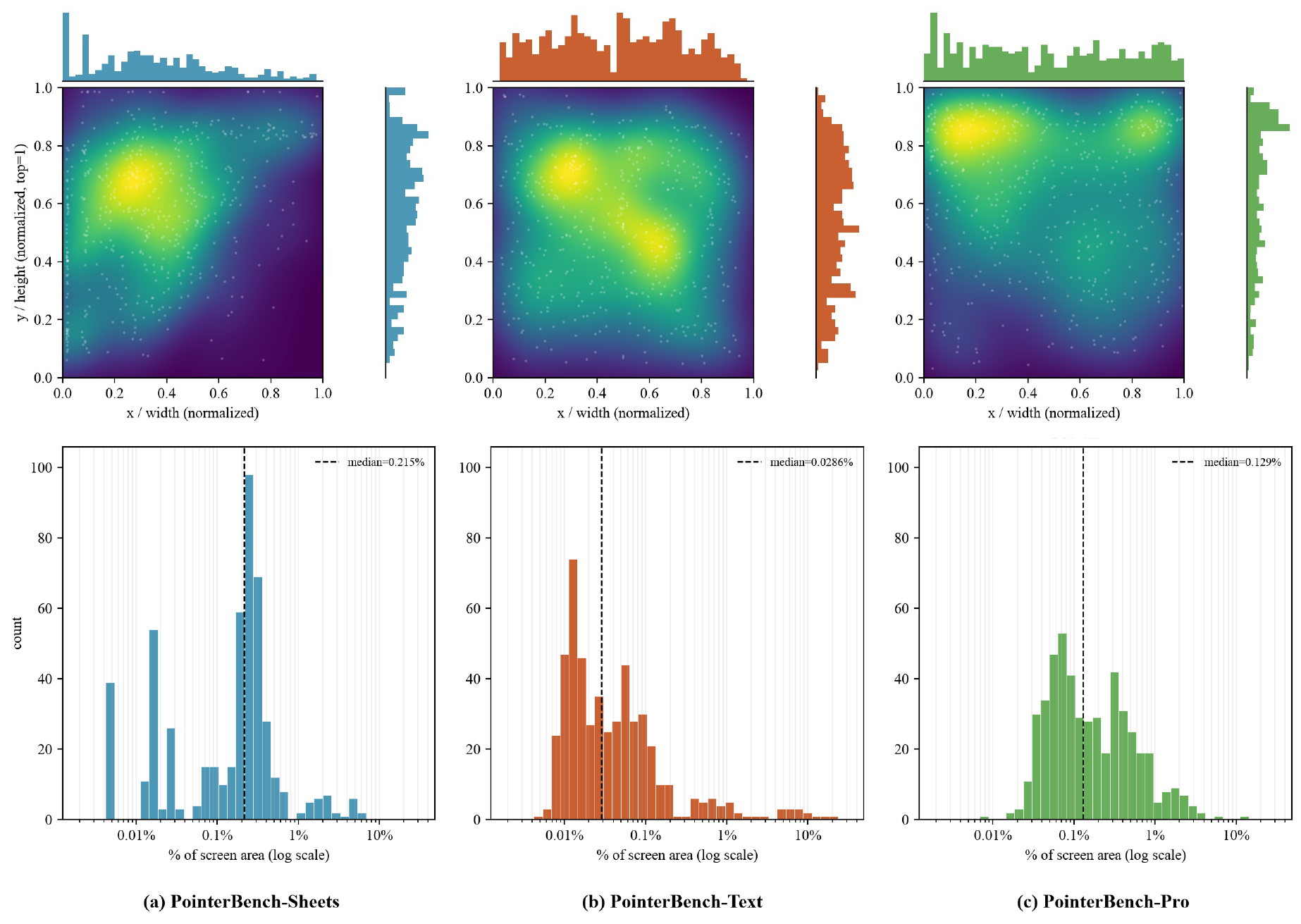}
    \caption{Spatial and geometric distributions of \pb{} targets. Top:
    normalized target-center distributions and their marginals. Bottom: target
    area as a percentage of the $1024\!\times\!768$ screen; dashed lines mark
    suite medians.}
    \label{fig:target-distributions}
\end{figure}

\vspace{-4mm}
\paragraph{Spatial and geometric coverage.}
Figure~\ref{fig:target-distributions} shows target-center distributions and
target areas. Sheets follows the regular geometry of spreadsheet layouts, Text
is spread across document surfaces, and Pro is concentrated toward menus,
toolbars, and other controls near the upper part of the interface. All three
suites nevertheless cover a broad portion of the coordinate frame. Median
target area is $0.215\%$ for Sheets, $0.0286\%$ for Text, and $0.129\%$ for
Pro; the median Text target is therefore approximately $7.5\times$ smaller
than the median Sheets target and $4.5\times$ smaller than the median Pro
target.

\begin{figure}[t]
    \centering
    \includegraphics[width=\linewidth]{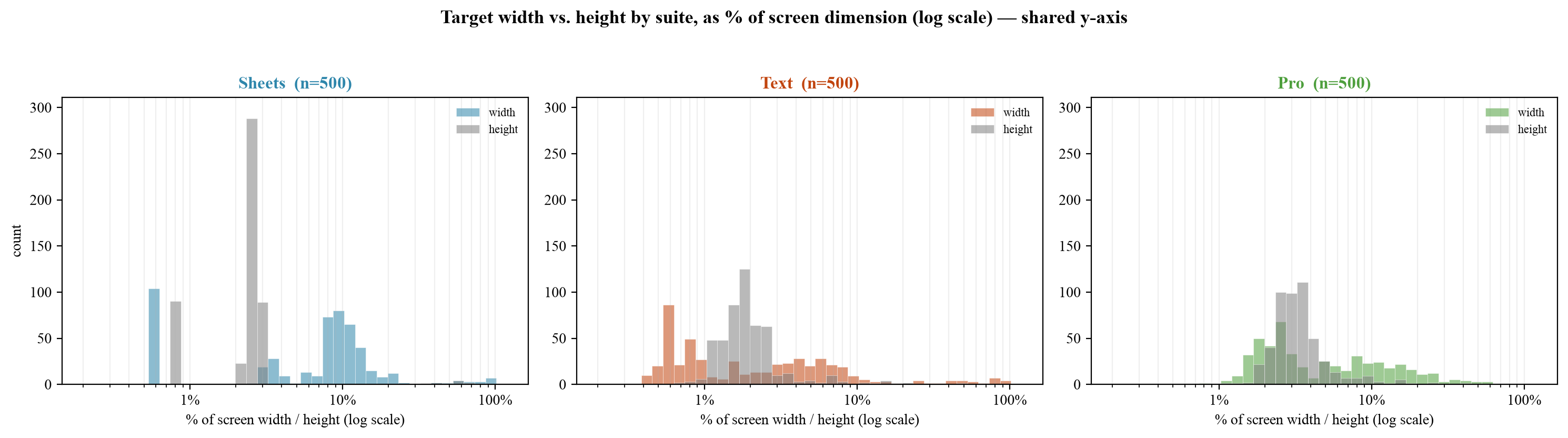}
    \caption{Normalized target width and height by suite, shown as percentages
    of the corresponding screen dimension on a logarithmic scale.}
    \label{fig:target-dimensions}
\end{figure}

Area alone can obscure highly anisotropic targets. A spreadsheet boundary, for
example, may cover a visible portion of the screen while remaining only a few
pixels wide. Figure~\ref{fig:target-dimensions} therefore reports target width
and height separately. Across the evaluated model--suite configurations, the
largest-area quartile is $10.7$ percentage points more accurate than the
smallest on average, although the size effect varies across suites and does not
fully explain the observed errors.

\vspace{-2mm}
\section{Evaluation Protocol and Inference Settings}
\label{app:evaluation}
\vspace{-2mm}
All models receive the same screenshot--instruction pairs and return either a
point or a bounding box in the benchmark coordinate frame. Point predictions
are correct when the predicted coordinate lies inside the annotated target.
For bounding-box tasks, let $B_p$ and $B_g$ denote the predicted and
annotated boxes. We compute
\begin{equation}
    \mathrm{coverage}=\frac{|B_p\cap B_g|}{|B_g|},
    \qquad
    \mathrm{precision}=\frac{|B_p\cap B_g|}{|B_p|},
\end{equation}
and count a prediction as correct when
$\mathrm{coverage}\geq0.90$ and $\mathrm{precision}\geq0.70$. The asymmetric
criterion strongly penalizes clipping the target while permitting a modest
margin around it. 

Malformed or unusable outputs are counted as failures. We report binary
success rate for each suite and their unweighted mean. Because predictions are
intended to be directly executable, official scores use the model's unmodified
output in the requested coordinate frame. We separately analyze models that
systematically use a different coordinate convention.

\paragraph{Prompt.}
All frontier model runs in the benchmark use the same system prompt, without
model-specific modifications:

\begin{center}
\colorbox{gray!10}{%
\begin{minipage}{0.94\linewidth}
\small\ttfamily
You are evaluating \pb{}, a GUI grounding benchmark. You will receive
one 1024x768 screenshot and one task instruction. Use absolute pixel
coordinates with origin at the top-left of the image. Do not return normalized
coordinates. Do not crop or resize the coordinate frame. For point tasks,
return JSON like \{"point": [x, y]\}. For bounding-box tasks, return JSON like
\{"bbox": [x0, y0, x1, y1]\}.
\end{minipage}%
}
\end{center}

\paragraph{Inference settings.}
All frontier model evaluations in the benchmark were completed through OpenRouter providers. Original $1024\!\times\!768$ images were
submitted directly, and the response format was enforced through a JSON schema.
Runs with reasoning disabled used a 512-token response cap. Low and medium
settings used a 2,048-token cap; medium additionally used a 1,024-token
reasoning allowance, while the low setting used the provider's low-effort
configuration without an explicit reasoning-token cap. Filtered or
budget-exhausted responses were not retried. Exact endpoint identifiers,
request manifests, hashes, and per-example outputs are included with the
released evaluation artifacts.

\begin{table}[h]
    \centering
    \small
    \caption{Reasoning settings available for the six models in the controlled
    reasoning study. A dash denotes an unavailable setting.}
    \label{tab:inference-settings}
    \begin{tabular}{lccc}
        \toprule
        Model & Disabled & Low & Medium \\
        \midrule
        Claude Fable 5 & -- & Yes & Yes \\
        Claude Sonnet 4.6 & Yes & Yes & Yes \\
        GPT-5.6 Luna & Yes & Yes & Yes \\
        GPT-5.6 Sol & Yes & Yes & Yes \\
        Grok 4.20 & Yes & -- & Yes \\
        Qwen3.6 Flash & Yes & -- & Yes \\
        \bottomrule
    \end{tabular}
\end{table}

\vspace{-4mm}
\section{Reasoning-Budget Sensitivity}
\label{app:reasoning}

We separately evaluate six models across their available reasoning settings
to measure sensitivity to inference budget on the same 1,500 examples.

\begin{table*}[h]
    \centering
    \small
    \caption{Success rate (\%) across available reasoning settings.}
    \label{tab:reasoning-sensitivity}
    \begin{tabular}{llrrrr}
        \toprule
        Model & Setting & Sheets & Text & Pro & Avg. \\
        \midrule
        Claude Fable 5 & Low & 95.8 & 67.4 & 82.2 & 81.8 \\
        Claude Fable 5 & Medium & 97.6 & 69.6 & 82.4 & 83.2 \\
        \addlinespace
        GPT-5.6 Sol & Disabled & 85.6 & 40.0 & 76.8 & 67.5 \\
        GPT-5.6 Sol & Low & 96.4 & 54.6 & 80.8 & 77.3 \\
        GPT-5.6 Sol & Medium & 97.8 & 53.4 & 82.0 & 77.7 \\
        \addlinespace
        GPT-5.6 Luna & Disabled & 58.8 & 27.2 & 68.8 & 51.6 \\
        GPT-5.6 Luna & Low & 87.6 & 45.4 & 79.4 & 70.8 \\
        GPT-5.6 Luna & Medium & 92.0 & 46.6 & 79.4 & 72.7 \\
        \addlinespace
        Claude Sonnet 4.6 & Disabled & 83.0 & 45.8 & 77.4 & 68.7 \\
        Claude Sonnet 4.6 & Low & 79.8 & 44.2 & 76.8 & 66.9 \\
        Claude Sonnet 4.6 & Medium & 88.8 & 49.0 & 78.8 & 72.2 \\
        \addlinespace
        Grok 4.20 & Disabled & 3.0 & 0.2 & 2.2 & 1.8 \\
        Grok 4.20 & Medium & 13.4 & 1.0 & 11.6 & 8.7 \\
        \addlinespace
        Qwen3.6 Flash & Disabled & 2.6 & 0.4 & 2.4 & 1.8 \\
        Qwen3.6 Flash & Medium & 2.6 & 0.0 & 2.0 & 1.5 \\
        \bottomrule
    \end{tabular}
\end{table*}

The effect is strongly model dependent. Luna improves from $51.6\%$ with
reasoning disabled to $70.8\%$ at low and $72.7\%$ at medium. Sol similarly
rises from $67.5\%$ to $77.3\%$ and $77.7\%$, making low reasoning a close
approximation to medium for this model. Sonnet is non-monotonic: its low setting
falls below the disabled run before medium reaches $72.2\%$. Fable has no
disabled run, but medium reasoning improves accuracy by only 1.4 percentage points relative to low reasoning. Medium reasoning
raises Grok from $1.8\%$ to $8.7\%$ while leaving it far behind the leading
models, and it does not improve Qwen. The main ranking is therefore not a
uniform consequence of allocating more reasoning tokens.

For Sonnet, Luna, and Sol, disabled and low runs are exactly paired at the
example level. Table~\ref{tab:reasoning-paired} reports the number of examples
corrected and regressed by the low setting. The changes are statistically
significant in all three cases, including Sonnet's small decline.

\begin{table}[h]
    \centering
    \small
    \caption{Paired comparison of low reasoning against reasoning disabled.
    Corrected denotes an incorrect-to-correct transition; regressed denotes the
    reverse. Reported $p$-values use two-sided exact McNemar tests.}
    \label{tab:reasoning-paired}
    \begin{tabular}{lrrrr}
        \toprule
        Model & $\Delta$ (pp) & Corrected & Regressed & $p$ \\
        \midrule
        Claude Sonnet 4.6 & $-1.80$ & 22 & 49 & $0.00182$ \\
        GPT-5.6 Luna & $+19.20$ & 337 & 49 & $5.59\!\times\!10^{-54}$ \\
        GPT-5.6 Sol & $+9.80$ & 199 & 52 & $1.90\!\times\!10^{-21}$ \\
        \bottomrule
    \end{tabular}
\end{table}

\vspace{-2mm}
\section{Additional Diagnostic Breakdowns}
\label{app:diagnostics}
\vspace{-2mm}
\paragraph{Spreadsheet target types.}
Table~\ref{tab:sheets-diagnostic} reports accuracy over the six coarse Sheets
groups. The leading systems are nearly perfect on cells, headers, colors, and
relative references, while thin structural targets remain more variable. Among
Fable, Opus, Sol, and Luna, edge accuracy ranges from $81.0\%$ to $98.3\%$
and corner accuracy from $89.7\%$ to $97.4\%$. Opus reaches $98.3\%$ on
edges and $94.9\%$ on corners, while Fable falls to $91.4\%$ on edges despite
perfect cell, header, and color grounding. These results confirm that high
aggregate Sheets accuracy can conceal residual errors on fine geometric targets.

\begin{table*}[h]
    \centering
    \small
    \caption{\pb{}-Sheets accuracy (\%) by target group. Counts are shown
    in the column headers.}
    \label{tab:sheets-diagnostic}
    \begin{tabular}{lrrrrrr}
        \toprule
        Model & Cell (170) & Header (59) & Color (56) & Relative (60) & Edge (116) & Corner (39) \\
        \midrule
        Claude Fable 5      & 100.0 & 100.0 & 100.0 & 98.3 & 91.4 & 97.4 \\
        Claude Opus 5       & 100.0 & 100.0 & 98.2  & 98.3 & 98.3 & 94.9 \\
        GPT-5.6 Sol         & 98.8  & 100.0 & 100.0 & 96.7 & 95.7 & 94.9 \\
        GPT-5.6 Luna        & 95.3  & 98.3  & 96.4  & 95.0 & 81.0 & 89.7 \\
        Grok 4.5            & 15.9  & 33.9  & 28.6  & 16.7 & 2.6  & 0.0 \\
        \bottomrule
    \end{tabular}
\end{table*}

\paragraph{Professional-interface target types.}
Table~\ref{tab:pro-diagnostic} confirms the ordering reported in the main
paper: text elements are easiest, other interface controls occupy an
intermediate position, and icons are hardest for every leading model. Across
Fable, Opus, Sol, and Luna, text accuracy ranges from $89.0\%$ to $92.9\%$,
compared with $81.4$--$85.9\%$ for other controls and only $70.0$--$75.3\%$
for icons. Icons account for 215 of 375 pooled failures ($57.3\%$) despite
representing only $38.0\%$ of the examples, confirming that icon grounding
remains the dominant source of error in professional interfaces.

\begin{table}[h]
    \centering
    \small
    \caption{\pb{}-Pro accuracy (\%) by target type.}
    \label{tab:pro-diagnostic}
    \begin{tabular}{lrrr}
        \toprule
        Model & Text (154) & Other controls (156) & Icon (190) \\
        \midrule
        Claude Fable 5      & 92.9 & 85.9 & 71.1 \\
        Claude Opus 5       & 92.2 & 83.3 & 70.5 \\
        GPT-5.6 Sol         & 90.9 & 81.4 & 75.3 \\
        GPT-5.6 Luna        & 89.0 & 81.4 & 70.0 \\
        Grok 4.5            & 20.8 & 16.0 & 6.8 \\
        \bottomrule
    \end{tabular}
\end{table}

\paragraph{Language.}
Table~\ref{tab:text-language} compares English and non-English Text examples
for the leading models. Multilinguality does not explain the Text bottleneck:
Fable performs identically on the two groups ($69.6\%$), Sol is nearly
unchanged ($53.2\%$ vs.\ $53.6\%$), and Luna is slightly stronger on
non-English examples ($45.6\%$ vs.\ $47.6\%$). Opus shows the largest difference, 
improving from 59.6\% on English to 66.8\% on non-English examples.

\begin{table}[h]
    \centering
    \small
    \caption{\pb{}-Text accuracy (\%) by language group. Each group contains
    250 examples.}
    \label{tab:text-language}
    \begin{tabular}{lrr}
        \toprule
        Model & English & Non-English \\
        \midrule
        Claude Fable 5      & 69.6 & 69.6 \\
        Claude Opus 5       & 59.6 & 66.8 \\
        GPT-5.6 Sol         & 53.2 & 53.6 \\
        GPT-5.6 Luna        & 45.6 & 47.6 \\
        Grok 4.5            & 1.6  & 1.2 \\
        \bottomrule
    \end{tabular}
\end{table}

\paragraph{Bounding-box failures.}
The Region group in the main paper contains 134 bounding-box examples.
Table~\ref{tab:bbox-failures} separates the resulting failures. Fable and
Opus most often fail by predicting boxes that are too large, with oversized
predictions accounting for $26.1\%$ of examples for both models. Sol and Luna
instead most often clip part of the requested region, at $39.6\%$ and
$41.8\%$, respectively. Thus, low Region accuracy does not arise from a
single shared failure mode even among the strongest models.

\begin{table*}[h]
    \centering
    \small
    \caption{Outcome distribution (\%) on the 134 \pb{}-Text bounding-box
    targets. Clipped and oversized boxes fail only the coverage or precision
    requirement, respectively; ``both'' fails both requirements.}
    \label{tab:bbox-failures}
    \begin{tabular}{lrrrrrr}
        \toprule
        Model & Correct & Clipped & Oversized & Both & Wrong region & Invalid \\
        \midrule
        Claude Fable 5      & 56.0 & 6.0  & 26.1 & 10.4 & 1.5  & 0.0 \\
        Claude Opus 5       & 50.7 & 12.7 & 26.1 & 7.5  & 3.0  & 0.0 \\
        GPT-5.6 Sol         & 30.6 & 39.6 & 15.7 & 9.0  & 3.0  & 2.2 \\
        GPT-5.6 Luna        & 26.1 & 41.8 & 9.7  & 10.4 & 8.2  & 3.7 \\
        Grok 4.5            & 0.7  & 3.0  & 3.7  & 18.7 & 73.9 & 0.0 \\
        \bottomrule
    \end{tabular}
\end{table*}

\paragraph{Coordinate-frame compliance.}
Gemini 3.7 Flash and Qwen3.8 Max predominantly emit coordinates in a
$0$--$1000$ frame despite the prompt explicitly requesting absolute
$1024\times768$ coordinates. We therefore omit their raw category-level
breakdowns from the preceding diagnostic analyses, where the coordinate
mismatch would confound grounding errors with output representation. Their
unmodified predictions remain in the official leaderboard. As a diagnostic,
uniformly converting parseable predictions to the requested frame raises
overall accuracy from $3.1\%$ to $78.1\%$ for Gemini and from $2.8\%$ to
$55.4\%$ for Qwen3.8 Max. Qwen3.6 Flash exhibits the same behavior, rising
from $1.5\%$ to $31.9\%$ under the same conversion. These transformed scores
are diagnostic only and are not used for ranking.

\begin{table}[h]
    \centering
    \small
    \caption{Coordinate-frame diagnostic. Official scores use unmodified model
    outputs; converted scores apply a fixed $0$--$1000$ to $1024\times768$
    transformation and are not leaderboard results.}
    \label{tab:coordinate-frame}
    \begin{tabular}{lrrr}
        \toprule
        Model & Official & Converted & $\Delta$ \\
        \midrule
        Gemini 3.7 Flash & 3.1 & 78.1 & +74.9 \\
        Qwen3.8 Max & 2.8 & 55.4 & +52.6 \\
        Qwen3.6 Flash (Medium) & 1.5 & 31.9 & +30.3 \\
        \bottomrule
    \end{tabular}
\end{table}

\section{Full Model Evaluation}
\label{app:full-results}

\begingroup
\small
\setlength{\LTcapwidth}{\textwidth}
\begin{longtable}{rllrrrr}
\caption{Full \textsc{PointerBench} evaluation. Success rate (\%) on the three benchmark suites. Newly evaluated Gemini and Qwen entries use official raw outputs without coordinate-frame post-processing.}
\label{tab:full-leaderboard}  \\
\toprule
Rank & Provider & Model & Sheets & Text & Pro & Avg. \\
\midrule
\endfirsthead

\multicolumn{7}{c}
{{\tablename\ \thetable{} -- continued from previous page}} \\
\toprule
Rank & Provider & Model & Sheets & Text & Pro & Avg. \\
\midrule
\endhead

\midrule
\multicolumn{7}{r}{\emph{Continued on next page}} \\
\endfoot

\bottomrule
\endlastfoot

1 & Anthropic & Claude Fable 5 & \underline{97.6} & \textbf{69.6} & \textbf{82.4} & \textbf{83.2} \\
2 & Anthropic & Claude Opus 5 & \textbf{98.8} & \underline{63.2} & 81.2 & \underline{81.1} \\
3 & OpenAI & GPT-5.6 Sol & 97.8 & 53.4 & \underline{82.0} & 77.7 \\
4 & OpenAI & GPT-5.6 Luna & 92.0 & 46.6 & 79.4 & 72.7 \\
5 & OpenAI & GPT-5.6 Terra & 91.6 & 47.6 & 77.6 & 72.3 \\
6 & Anthropic & Claude Sonnet 4.6 & 88.8 & 49.0 & 78.8 & 72.2 \\
7 & Anthropic & Claude Opus 4.8 & 82.2 & 44.4 & 80.2 & 68.9 \\
8 & OpenAI & GPT-5.6 Sol Pro & 92.0 & 32.0 & 72.0 & 65.3 \\
9 & Anthropic & Claude Sonnet 5 & 78.6 & 36.0 & 77.0 & 63.9 \\
10 & Anthropic & Claude Opus 4.7 & 73.8 & 36.2 & 75.4 & 61.8 \\
11 & OpenAI & GPT-5.5 & 72.4 & 31.4 & 75.8 & 59.9 \\
12 & OpenAI & GPT-5.6 Terra Pro & 52.0 & 20.0 & 74.0 & 48.7 \\
13 & OpenAI & GPT-5.4 & 63.2 & 25.0 & 50.4 & 46.2 \\
14 & OpenAI & GPT-5.6 Luna Pro & 54.0 & 18.0 & 62.0 & 44.7 \\
15 & OpenAI & GPT-5 & 35.2 & 13.6 & 20.2 & 23.0 \\
16 & xAI & Grok 4.5 & 15.2 & 1.4 & 14.0 & 10.2 \\
17 & xAI & Grok 4.20 & 13.4 & 1.0 & 11.6 & 8.7 \\
18 & Moonshot & Kimi K2.5 & 10.8 & 3.4 & 3.0 & 5.7 \\
19 & OpenAI & GPT-5 Mini & 6.0 & 1.4 & 5.8 & 4.4 \\
20 & Moonshot & Kimi K2.6 & 5.6 & 2.8 & 2.8 & 3.7 \\
21 & Qwen & Qwen3.7 Plus & 5.8 & 0.6 & 3.0 & 3.1 \\
22 & Google & Gemini 3.7 Flash & 4.0 & 1.2 & 4.2 & 3.1 \\
23 & Qwen & Qwen3.8 Max & 4.8 & 0.2 & 3.4 & 2.8 \\
24 & MiniMax & MiniMax M3 & 3.6 & 0.4 & 3.6 & 2.5 \\
25 & Qwen & Qwen3.6 Plus & 3.0 & 0.6 & 3.6 & 2.4 \\
26 & Qwen & \shortstack[l]{Qwen3-VL-235B-A22B\\Thinking} & 2.6 & 0.4 & 3.2 & 2.1 \\
27 & Qwen & Qwen3.5-9B & 2.0 & 0.2 & 2.6 & 1.6 \\
28 & Google & Gemini 3.5 Flash & 3.4 & 0.6 & 0.6 & 1.5 \\
29 & Qwen & Qwen3.6 Flash & 2.6 & 0.0 & 2.0 & 1.5 \\
30 & OpenAI & GPT-5 Nano & 2.0 & 0.2 & 1.6 & 1.3 \\
31 & Moonshot & Kimi K2.7 Code & 0.4 & 1.8 & 0.8 & 1.0 \\
32 & xAI & Grok 4.3 & 1.2 & 0.0 & 0.0 & 0.4 \\
33 & Google & Gemini 3.1 Pro Preview & 0.0 & 0.2 & 0.4 & 0.2 \\
34 & Google & \shortstack[l]{Gemini 3.1 Pro Preview\\(Custom Tools)} & 0.0 & 0.4 & 0.0 & 0.1 \\
35 & Stepfun & Step-3.7-Flash & 0.0 & 0.0 & 0.2 & 0.1 \\
36 & xAI & Grok Build 0.1 & 0.0 & 0.0 & 0.0 & 0.0 \\
37 & Xiaomi & MiMo V2.5 & 0.0 & 0.0 & 0.0 & 0.0 \\

\midrule
-- & Baseline & Center click & 0.4 & 0.2 & 0.2 & 0.3 \\

\end{longtable}
\endgroup

Table~\ref{tab:full-leaderboard} contains all 37 evaluated systems together
with the center-click baseline.

\section{Complete Specialist GUI Grounding Evaluation}
\label{app:specialists}

Table~\ref{tab:specialist-grounding} reports the complete evaluation of 39 open-weight specialist GUI grounding configurations (spanning 35 checkpoints from 2B to 32B parameters) across all 1,500 \pb{} examples. Unlike Table~\ref{tab:full-leaderboard}, which tests strict adherence to the unified absolute-pixel JSON prompt, each specialist model here is evaluated under its official native grounding prompt, coordinate normalization (e.g., $[0,1000]$ normalized coordinates for UI-Venus-2-9B and UI-Venus-1.5-8B, $[0,999]$ scaled coordinates for MAI-UI-8B and MAI-UI-2B, Doubao grounding format for UI-TARS-1.5-7B, and \texttt{pyautogui.click} for GTA1 and OpenCUA), or inference-time scaling wrapper (MVP and GUI-RC~\citep{du2026guirc}). On region tasks, a returned box is scored as a box and a point-only output is scored as a point.

\input{tables/specialist_models}

\section{Additional Limitations}
\label{app:limitations}

\pb{} isolates one-step GUI grounding from temporal planning, recovery, and
environment dynamics. Its scores should therefore not be interpreted as a
complete measure of computer-use-agent capability; conversely, success on an
interactive benchmark does not guarantee robustness to the fine-grained
targets considered here.
All inputs use a fixed $1024\!\times\!768$ frame and absolute pixel
coordinates. The benchmark does not directly measure robustness to arbitrary
display resolutions, resizing policies, or alternative coordinate
representations.

As with other public benchmarks evaluated through closed models, we cannot
independently verify whether benchmark examples or closely related material
have appeared in proprietary pretraining corpora. \pb{}-Sheets and
\pb{}-Text are generated specifically for the benchmark, while
\pb{}-Pro uses generated professional interfaces, reducing direct
overlap with established GUI grounding datasets but not eliminating the
possibility of exposure to similar interfaces or tasks.

%% file: tables/specialist_models.tex
\begingroup
\small
\setlength{\LTcapwidth}{\textwidth}
\begin{longtable}{rllrrrr}
\caption{Specialist GUI grounding results on \pb{}. Success rate (\%) on the
three suites. Best and second-best results in each column are bold and underlined.}
\label{tab:specialist-grounding} \\
\toprule
Rank & Family & Model & Sheets & Text & Pro & Avg. \\
\midrule
\endfirsthead
\multicolumn{7}{c}
{{\tablename\ \thetable{} -- continued from previous page}} \\
\toprule
Rank & Family & Model & Sheets & Text & Pro & Avg. \\
\midrule
\endhead
\midrule
\multicolumn{7}{r}{\emph{Continued on next page}} \\
\endfoot
\bottomrule
\endlastfoot
1  & MVP & Qwen3-VL-32B & \textbf{68.4} & \textbf{34.0} & \textbf{75.4} & \textbf{59.3} \\
2  & GTA1 & 32B & 65.0 & \underline{29.6} & 70.2 & \underline{54.9} \\
3  & UI-Venus & 2-9B & 64.2 & 25.8 & \underline{73.6} & 54.5 \\
4  & MAI-UI & 8B & 59.2 & 23.2 & 71.6 & 51.3 \\
5  & MVP & Qwen3-VL-8B & 60.4 & 22.4 & 70.8 & 51.2 \\
6  & UI-Venus & 1.5-8B & 56.4 & 23.8 & 72.8 & 51.0 \\
7  & OpenCUA & 32B & \underline{66.0} & 19.4 & 67.4 & 50.9 \\
8  & Qwen3-VL & 32B-Instruct & 58.8 & 23.8 & 69.6 & 50.7 \\
9  & ScaleCUA & 32B & 57.4 & 27.4 & 66.8 & 50.5 \\
10 & GTA1 & 7B & 60.2 & 22.8 & 67.4 & 50.1 \\
11 & OpenCUA & 7B & 59.8 & 18.4 & 63.8 & 47.3 \\
12 & MVP & UI-TARS-1.5-7B & 50.6 & 26.8 & 62.6 & 46.7 \\
13 & UI-Venus & Ground-7B & 47.8 & 24.6 & 66.2 & 46.2 \\
14 & ScaleCUA & 7B & 51.6 & 22.0 & 63.6 & 45.7 \\
15 & GUI-G2 & 7B & 45.8 & 21.6 & 64.6 & 44.0 \\
16 & UI-Ins & 32B & 44.0 & 24.0 & 62.8 & 43.6 \\
17 & Qwen3-VL & 8B-Instruct & 48.2 & 14.6 & 65.8 & 42.9 \\
18 & UI-Ins & 7B & 42.6 & 22.0 & 62.2 & 42.3 \\
19 & Jedi & 7B-1080p & 47.4 & 20.4 & 57.2 & 41.7 \\
20 & UI-TARS & 1.5-7B & 46.2 & 18.2 & 57.4 & 40.6 \\
21 & InfiGUI & G1-7B & 41.0 & 18.6 & 59.2 & 39.6 \\
22 & SE-GUI & 7B & 40.2 & 20.2 & 53.4 & 37.9 \\
23 & GUI-RC & UI-TARS-1.5-7B & 45.6 & 15.6 & 51.4 & 37.5 \\
24 & ScaleCUA & 3B & 36.4 & 17.2 & 58.0 & 37.2 \\
25 & Jedi & 3B-1080p & 37.0 & 16.8 & 57.6 & 37.1 \\
26 & Rex-Omni & 3B & 43.6 & 17.6 & 49.4 & 36.9 \\
27 & MAI-UI & 2B & 36.6 & 9.8 & 61.0 & 35.8 \\
28 & GUI-G2 & 3B & 29.6 & 19.2 & 57.8 & 35.5 \\
29 & FocusUI & 7B & 32.8 & 12.4 & 56.0 & 33.7 \\
30 & UGround & V1-7B & 36.8 & 13.6 & 50.6 & 33.7 \\
31 & GUI-Actor & 7B & 35.6 & 11.0 & 53.6 & 33.4 \\
32 & GUI-Spotlight & 7B & 34.4 & 11.8 & 53.6 & 33.3 \\
33 & GUI-AIMA & 3B & 33.6 & 9.2 & 56.8 & 33.2 \\
34 & GUI-Actor & 3B & 32.6 & 11.4 & 50.6 & 31.5 \\
35 & InfiGUI & G1-3B & 23.2 & 15.2 & 53.6 & 30.7 \\
36 & FocusUI & 3B & 27.0 & 11.2 & 51.8 & 30.0 \\
37 & FocusUI & Qwen3-VL-2B & 27.8 & 9.0 & 51.0 & 29.3 \\
38 & UGround & V1-2B & 27.0 & 9.6 & 45.0 & 27.2 \\
39 & SE-GUI & 3B & 19.2 & 7.8 & 48.8 & 25.3 \\
\end{longtable}
\endgroup